\documentclass[conference]{IEEEtran}
\usepackage[utf8]{inputenc}
\usepackage{lipsum} 
\usepackage{epsfig}
\usepackage{float}
\usepackage{bbold}
\usepackage{bm}
\usepackage{hyperref}
\usepackage{cite}
\usepackage{amsmath,amssymb}
\usepackage{graphicx}
\usepackage{hyperref}
\usepackage{xcolor}
\usepackage{stmaryrd}
\usepackage{hhline} 
\usepackage{makecell} 
\usepackage{pifont}  
\usepackage{url} 
\usepackage{cuted}  
\usepackage[labelfont=bf,labelsep=period]{caption}

\usepackage{color,soul} 
\usepackage[english]{babel}
\usepackage{amsmath}

\date{July 2025}

\begin{document}

\begin{strip}
\begin{center}
    {\fontsize{16pt}{22pt}\selectfont \bfseries Towards Robust Prehensile Manipulation in Open-Ended Environments \par}
    \vspace{0.5em}
    {\large
        Mathilde Kappel,
        Mahdi Khoramshahi,
        Louis Annabi,
        Faïz Ben Amar,
        Stéphane Doncieux \par
    }
    \vspace{0.2em}
    {\small
    Sorbonne Université, CNRS, Institut des Systèmes Intelligents et de Robotique (ISIR), Paris, France \\
    \texttt{\{mathilde.kappel, mahdi.khoramshahi, louis.annabi, faiz.benamar, stephane.doncieux\}@isir.upmc.fr}
    }
\end{center}

\vspace{1em}

\begin{abstract}
We propose to use Quality-Diversity (QD) algorithms to solve robotic prehensile manipulation tasks in open-ended environments. Our approach enables the efficient discovery of a wide range robust grasp configurations, which serve as reliable starting points for generating diverse prehensile manipulation trajectories on articulated objects. The resulting diversity in manipulation behaviors enhances generalization and adaptability, enabling effective deployment continuously evolving open-world settings.
\end{abstract}

\vspace{1em}

\begin{center}
    \includegraphics[width=0.8\textwidth]{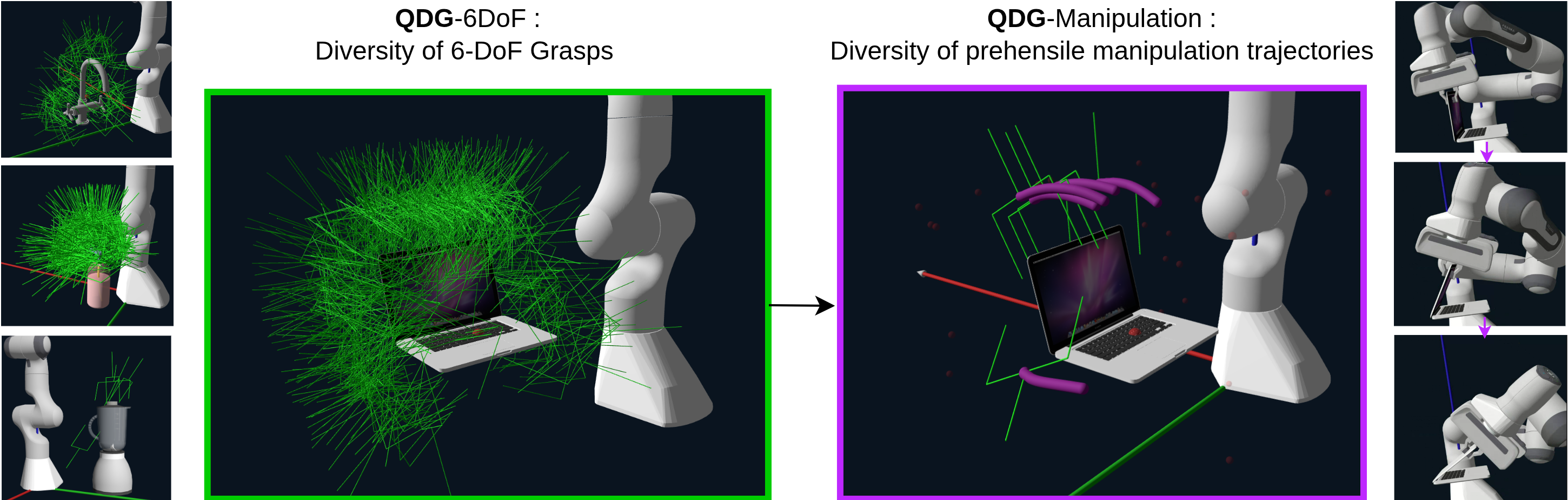}
    \captionsetup[figure]{name=Fig.}
    \captionof{figure}{From Quality-Diversity grasp generation to prehensile manipulation: the QDG-Manipulation pipeline.}
    \label{fig:my_label}
\end{center}
\end{strip}

\vspace{1em}

\section*{1. Introduction to Prehensile Manipulation}
Developing algorithms enabling robot prehensile manipulation tasks in dynamic real-world environments remains a major challenge in robotics. The manipulation problem aims at finding a set of suitable robot controls to change the state of an object from an initial to a desired configuration. More specifically, prehensile manipulation involves keeping the object grasped while the whole robot executes the manipulation using its control system  \cite{morlans2025ao}. For example opening a fridge is prehensile manipulating switching the object state from closed to open by grasping the fridge hammer during the whole robot motion. 
Solving robotic prehensile manipulation problems is inherently difficult due to the sparse nature of the grasping problem. In such scenarios, the space of solutions is extremely large, yet successful feedback is rare, making learning-based approaches particularly challenging. Moreover, the environments in which robots operate are highly variable—each static scene can differ significantly in object positions, shapes, levels of clutter, and semantic context \cite{appius2025taskaware}. For these reasons, prehensile manipulation is often described as an \textit{open-world} or \textit{open-ended} problem. 

One promising approach to tackle the complexity of open-word environment problems is using real-time adaptation to the environment based on spontaneous exploration. Curiosity-driven learning and intrinsic motivation methods exist to solve sparse reward tasks. More specifically, Quality-Diversity (QD) algorithms have been designed to efficiently explore large and sparse environments by generating a diverse set of high-quality solutions, rather than converging on a single optimal policy. Precisely, Quality-Diversity algorithms are evolutionary algorithms.  Evolutionary algorithms iteratively apply variation, evaluation and selection processes to a population of candidate solutions. QD methods are evolutionary algorithms that additionally optimize the diversity of solutions in the population according to behavior descriptors designed by the user \cite{mouret2015illuminating}.

The objective of our research is to use the Quality-Diversity open-ended learning method \cite{huber2023quality} to solve a real-world application: prehensile grasp manipulation. First, we generate a diverse repertoire of prehensile grasps, then we use these grasps to generate articulated object manipulation trajectories.

\section*{2. QDG-Manipulation}

\vspace{0.1cm}
\textbf{QDG-6DoF grasp module \cite{huber2024speeding}}: The QDG-6DoF module takes as input a target object and an end-effector model, and produces as output an archive of 6-DoF grasp poses distributed around the object. This module is built upon the \textit{Quality-Diversity (QD)} algorithm, in which the core components—namely the \textit{genome}, \textit{population}, \textit{behavior}, and \textit{fitness}—are specifically adapted to address the challenge of grasping in robotics. In this context, each grasp candidate is treated as an \textit{individual} within the population. The \textit{genome} of an individual encodes the Cartesian position of the end-effector $(x, y, z) \in \mathbb{R}^3$, its orientation represented as a unit quaternion $(q_1, q_2, q_3, q_4)$, and a synergy component from a predefined synergy space. These parameters define a continuous 6-DoF grasp configuration for the end-effector. The \textit{behavior space} is designed to reflect the structure of the 3D environment surrounding the articulated object. Each grasp candidate is associated with a \textit{behavior descriptor}, defined as the voxel in a discretized 3D grid that best approximates the spatial location of the grasp. A deterministic behavior mapping function assigns each individual to a unique cell in this space, effectively partitioning the continuous grasp space into manageable, behaviorally distinct regions. The \textit{fitness} of an individual corresponds to a success score that evaluates the quality of the grasp, based on whether the proposed configuration achieves a stable and feasible grasp under physical simulation. The objective of the QD module is to construct a MAP-Elites structured archive, in which each cell in the discretized behavior space is populated with the highest-performing grasp candidate among all those projected to that cell. This process relies on \textit{local competition}, where individuals only compete with others sharing the same behavior descriptor. Such a mechanism promotes both grasp diversity across space and optimality within each spatial region. To efficiently explore the grasp behavior space, the module applies a \textit{mutation-selection} process that iteratively refines the active population. Individuals are sampled from the archive and mutated to generate new grasp candidates, which are then evaluated and reinserted based on their fitness and behavior descriptor. The search is further guided by physical priors, allowing the projection from the search space to the grasp space to be grounded in physically plausible configurations.

\vspace{0.1cm}
\textbf{Bootstrap method to speed up grasp generation \cite{huber2024qdgset}: }
In order to scale up the generation of synthetic grasps using QDG-6DoF, we combine object mesh augmentation with transfer learning from previously acquired grasp repertoires. Starting with a dataset of grasp poses defined over a set of core objects $\textsc{P} = (\mathcal{P}_i)_{1 \leq i \leq n}$, where $n$ denotes the number of original objects, we apply a mesh augmentation process to each object. This yields an extended dataset $\textsc{P}' = (\mathcal{P}_{i}^{a})_{1 \leq i \leq n,\ 1 \leq a \leq a_{\text{max}}}$, where $a_{\text{max}}$ is the number of data augmentations per object. The resulting augmented objects are then provided with a QDG-6DoF
process, which is initialized with the grasps found on the
reference object.
 
\vspace{0.1cm}
\textbf{From grasp to articulated object manipulation}: When applied to articulated objects, the QD-Grasp-6DoF module generates a repertoire of diverse and successful grasps across different object parts. These grasps are not only stable but also enable physical interaction. After establishing contact, the robot enters a compliant mode and passively follows the motion induced by the object's articulation. During execution, the robot switches to active control to replay the trajectory and manipulate the object accordingly. Thus, each grasp is associated with both a stable contact and a dynamic manipulation behavior (Fig. \ref{fig:my_label}).

\section*{3. Results}
Using Quality-Diversity (QD) methods with priors significantly accelerates 6-DoF grasp sampling for prehensile manipulation and enables the construction of large datasets containing diverse and robust grasps. The QDG-6DoF module is compatible with various robotic platforms, ranging from simple parallel 2-finger grippers to highly dexterous hands. Furthermore, the QDG-6DoF framework was extended with a bootstrap method applied to augmented data. As a result, we introduce QDGset, a prehensile grasp dataset comprising 62 million grasps across 5,000 distinct objects—3.5 times larger than the widely used ACRONYM dataset \cite{eppner2020acronym}.

Beyond grasping, QDGset serves as a rich source of initial configurations for generating diverse manipulation trajectories. The diversity of these trajectories is a direct consequence of the underlying grasp variability provided by QDGset. 

\section*{4 . Conclusion and Future Work}
This study demonstrates that Quality-Diversity algorithms, originally developed for exploration in sparse and high-dimensional environments, can be effectively applied to solve prehensile manipulation tasks in open-world robotics. By leveraging QD’s capacity to generate a repertoire of diverse and high-quality grasps, we overcome the sparse reward challenge inherent to grasping problems. The diversity is essential in open-domain tasks where a robot must manipulate objects under changing conditions.

We are currently working on incorporating compliant control mechanisms into the way the prehensile manipulation grasps are executed by the robot. The robot should respond intelligently to physical interactions while following the trajectories provided by the Quality-Diversity trajectory repertoire. Such mechanisms not only enhance adaptability and robustness in the face of real-world uncertainties but also have the potential to reduce the sim-to-real gap.

\end{document}